# Content Based Image Retrieval (CBIR) in Remote Clinical Diagnosis and Healthcare

Albany E. Herrmann
*University of Tuebingen, Germany*

Vania V. Estrela
*Universidade Federal Fluminense, Brazil*

**ABSTRACT**
*Content-Based Image Retrieval (CBIR) locates, retrieves and displays images alike to one given as a query, using a set of features. It demands accessible data in medical archives and from medical equipment, to infer meaning after some processing. A problem similar in some sense to the target image can aid clinicians. CBIR complements text-based retrieval and improves evidence-based diagnosis, administration, teaching, and research in healthcare. It facilitates visual/automatic diagnosis and decision-making in real-time remote consultation/screening, store-and-forward tests, home care assistance and overall patient surveillance. Metrics help comparing visual data and improve diagnostic. Specially designed architectures can benefit from the application scenario. CBIR use calls for file storage standardization, querying procedures, efficient image transmission, realistic databases, global availability, access simplicity, and Internet-based structures. This chapter recommends important and complex aspects required to handle visual content in healthcare.*

## 1 INTRODUCTION

Medical Image Processing (MIP) is a major ally in e-health and telemedicine, enabling rapid diagnosis with visual, quantitative and analytical assessment (Fessler, 2009). Remote care can reveal subtle changes that indicate the progression of a therapy. Health facilities now have images from plenty of sources which leads to multidimensional images (2D, 3D, 4D and time-varying, to name a few possibilities), and multimodality images. For instance, Alzheimer's disease evaluation still uses behavioral and cognitive tests along with MRI and PET scans of the entire brain (Datta et al., 2008; Deselaers, 2014; Lew et al., 2006). Diverse image collections offer the chance to improve evidence-based diagnosis, administration, teaching, and research. There is a necessity for proper methods to search those collections for images that have similarity in some sense. Statistical bias can be reduced as discoveries are assessed without direct patient contact like quicker and more objective assessment of the effects of anticancer drugs.

Content-Based Image Retrieval (CBIR) is an image search framework that complements the usual text-based retrieval of images through visual features, such as color, shape, and texture as search criteria.

CBIR can be applied to multidimensional image retrieval, multimodality health data, and the recuperation of unusual datasets.

CBIR systems (CBIRSs) can be divided into two classes: Narrow Domain Applications (NDA) and Broad Domain Applications (BDA).

Medical Imagery Retrieval, Finger Print Retrieval, and Satellite Imagery Retrieval are types of NDA. These applications have small variance of content; target specific sources of knowledge; have homogeneous semantics; are likely to have some sort of ground truth; their content description is more objective; may have some control of scenes and sensors; involve limited interactivity; use quantitative evaluation; have tailored/data-driven architectures; are medium-sized; use object recognition techniques most of the times; and consider specific invariances with model-drivel tools.

Photo collections and Internet content are examples of BDA. They have high content variance; tackle generic sources of knowledge; have heterogeneous semantics; do not have any sort of ground truth; their content description is more subjective; no control of scenes and sensors; involve pervasiveness and interactivity; use qualitative evaluation; have modular/iteration-driven architectures; range from large to extremely large sizes; use information retrieval techniques most of the times; consider tools that are perceptual/cultural, in addition to general invariants.

Besides facilitating visual/automatic diagnosis and decision making, images can help real-time remote consultation and screening, store-and-forward examinations, home care assistance and overall patient surveillance. These applications involve a great deal of data that require web-based and other telemedicine structures. When it comes to the use of information systems in healthcare, some ideas must be clarified. Next, some relevant definitions are given (Oh, et al., 2005; Embi & Payne, 2009):

(i) **Hospital Information System (HIS)** is a segment of health informatics or e-Health that consists of a broad integrated data system designed to handle all the aspects related to electronic/digital processes in health (local and global); administrative, financial, legal and the corresponding service processing needs and practices in healthcare using the Internet as well as other types of communication settings. An important aspect a today's HIS is the ability of integrating home care, small clinics (local care) and big hospitals/research centers (global care), in order to better handle all patient information sources.

(ii) A **Picture Archiving and Communication System (PACS)** allows for inexpensive storage of and suitable access to several kinds of medical imaging (graphs, still pictures, video, streaming, etc.). This helps eliminating the need to perform manually some tasks and speeds up electronic images handling, and reports can be transmitted digitally via PACS (Oosterwijk, 2004). A PACS consists of the part that acquires images; a safe network to broadcast patient records; viewing/display hardware to help decision makers; images suitable for interpretation, annotation, plus reviewing; and adequate file storage and retrieval (where these archives can be images, reports or both, as well as some alternative sort of information fusion). If available and emerging web technologies are merged, then PACS can aid the access to images, interpretations, and related data, destroying the physical and time barriers related to conventional image retrieval, handling, distribution, and display. If possible, a complete PACS should offer a single point of access to images, their related information, and it should support all digital varieties during the phases of processing.

(iii) A **Radiology Information System (RIS)** is an automated database developed for radiology users and facilities to accumulate, work with, and distribute patient radiological information and visual descriptions. Its framework habitually consists of a means to track and schedule patients, report results, and image analysis capabilities. RIS complements HIS, and it is critical for an efficient workflow of radiology procedures (Mueller et al., 2004; Constantinescu et al., 2009).

(iv) **Electronic Health Records (EHRs)** (Tang et al., 2006; Redling, 2012) enable the exchange of patient data between different healthcare professionals (groups, specialists, etc.). The terms **Medical Record** and **Health Record** are used rather interchangeably to portray the systematic records of a single patient's medical past and care across time within one particular healthcare provider's place. Health records include several types of notes entered over time by healthcare professionals, recording observations and administration of medicines and treatments, recommendations for the administration of remedies and therapies, test outcomes, X-rays, reports, and so on. The preservation of full and precise medical records is a prerequisite of healthcare providers and it is, in general, enforced by a license or certification requirement.

(v) A **Computerized Physician Order Entry (CPOE)** refers to any arrangement in which clinicians directly and, progressively add in medication orders, tests and procedures toward a computer system to release the order straightforwardly to a pharmacy later (Lindenauer et al., 2006).

(vi) **ePrescribing** involves the access to prescriptions possibilities, printing given prescriptions to patients and electronic diffusion of prescriptions from physicians to pharmacists.

(vii) **Pharmacoinformatics (PI)** (also called **Pharmacy Informatics**) focuses on medication-related information and knowledge in a variety of healthcare systems (Troiano, 1999; Holler, 2013; Goldmann et al., 2014). It uses computers to store, retrieve and analyze drug and prescription data. PI management systems help pharmacists to make safe decisions about patient drugs/therapies, medical insurance records, drug interactions, evaluation of a patient's data, the modeling/simulation of drugs behavior, and the control of their performance by individualized dosage regimens for each patient to attain unambiguously chosen therapeutic goals.

(viii) **Telemedicine** comprises the physical and the psychological diagnosis and treatments at a distance, including surveillance of patients' functions.

(ix) **Consumer Health Informatics** is the use of electronic resources on health issues by sound individuals or patients.

(x) **Health Knowledge Management:** It is a structure that allows an overview of the latest medical literature, best practice guidelines or epidemiological trackings such as Medscape, PubMed Central (PMC), SciELO and MDLinx.

(xi) **Virtual Healthcare Teams** encompass the set of healthcare experts responsible for pooling resources and sharing data on patients through digital equipment and environments (Greengard, 2013).

(xii) **mHealth or m-Health** uses mobile devices to collect both collective and patient's health data, offering healthcare information to practitioners, nurses, researchers, and patients, with real-time monitoring of the patient's vitals, and straight provision of care.

(xiii) **Health Research with Grids** is a potent manner of computing and handling data administration capabilities that involve large amounts of heterogeneous data (Greengard, 2013).

(xiv) A **Nuclear Medicine Information Management System** (**NMIS**) deals with computational issues surrounding nuclear medicine use in healthcare facilities. Receptionists plan patient studies, register patient data, and produce examination records. Administrative personnel write down and print medical reports, find patient films, follow borrowed studies, access statistical reports, billing information, and resource use information. Technologists generate study worksheets, document quality assurance data and process the examination results. Radiopharmaceuticals specialists use the system to gather/record tracer stocks, to follow preparation and provision of activities, and to retain records for external evaluations. Doctors employ it to seek old results and document their analytical insights. Software applications may include patient scheduling, radiopharmacy, film management, report generation, quality reassurance, and inventory organization where implementations involve a diversity of educational and commercial systems. Hardware, its specification and installation issues are also considered.

(xv) A **Laboratory Information System (LIS)** is a software-based laboratory and information management system with features that support advanced laboratory operations. Central features take account of workflow, data handling support, data exchange interfaces, and flexible architecture, which fully support its use in regulated environments (Skobolev, et al., 2011). The features and applications of an LIS have evolved along time from uncomplicated sample tracking to an Enterprise Resource Planning (ERP) tool to manage multiple aspects of laboratory information processing. LISs are dynamic, since the laboratory constraints are rapidly evolving, and dissimilar labs often have different needs. Therefore, a working definition of an LIS depends on the comprehension by the individuals or groups involved. Recently, LIS functionality has extended beyond its original purpose of sample management. Other concerns are data management, data mining, data analysis, and Electronic Laboratory Notebook (ELN) integration. These features have been added to many LISs, making translational medicine possible entirely by a single software solution.

(xvi) **Clinical Decision Support** provides data electronically about protocols and standards for healthcare experts to employ in analyzing and treating patients.

The concepts explored above are far from being complete, but they signal the importance and complexity to handle visual content in healthcare.

## 2 IMAGE AND INFORMATION PROCESSING FUNDAMENTALS

An image is a 2D signal containing information about the image brightnesses and it can be represented as a function $\mathbf{f}(x, y)$, where the coordinates $(x, y)$ represent the spatial location of a pixel (Gonzalez & Woods, 2004). The value of $\mathbf{f}(x, y)$ is also called gray level or image intensity. Images can be of two types: continuous and discrete.

A continuous image is a real function of two continuous variables, such as the intensity of a photograph recorded on a film, is a 2D real function $\mathbf{f}(x, y)$ of two real-valued variables $x$ and y.

In eHealth and telemedicine, the simplest type of image is a discrete 2D function $\mathbf{I}(i, j)$ of two integer-valued variables $i$ and $j$, as in a scanned and discretized 320×240 pixel photograph. Thus, $\mathbf{f}$ becomes a 2D matrix $\mathbf{I}$ of size 320×240. A color image is habitually represented by three discrete matrices. For instance, a 320×240 pixel color picture can be split into three 320×240 channels (matrices), as follows: Red (R), Green (G), and Blue (B).

The next point is to pin down the notion of **image query**. In essence, it consists of the visual content a potential client is looking for.

The subsequent difficulty is how the user will retrieve relevant multimedia material. **Content** might refer to colors, shapes, categories resembling the query, textures, or any other pertinent clue that can result from consulting immense image collections on the basis of syntactical image features.

The term **metadata** is often used as an allusion to what is known about the items contained in a large database. For example, who inserted the data and in what format they are. Hence, metadata may refer to the knowledge that describes the context of another value. The words type, attribute, object, property, aspect, and schema all refer to metadata in some context. When only words are used, there is room for inconsistencies because of the variation of human interpretation.

In computer science, a **schema** corresponds to the organization or structure for a database. The process of data modeling leads to a schema (whose plural is schemata). The term is used in both relational and object-oriented databases. The word sometimes seems to refer to the visualization of a structure and, at times, to a formal text-oriented description.

A **descriptor** is a piece of stored data used to identify an item in a data storage and retrieval system. Descriptors keep pertinent information on content, saving the processing time in future queries by using image features that are essential for search and comparison.

**Semantic gap** is the difference between the human perception of a concept and how it can be represented in computers. An important aspect is to focus on designing image features to decrease the semantic gap involving visual features and the richness of human semantics.

The choice of standards can affect CBIR, even though these effects may well turn out to be minimal in practice. Three main areas of potential impact can be identified: image compression, query specification, and metadata description. Several CBIR aspects need to be taken into account. They include:

  i.  Network protocols such as TCP/IP, governing the transmission of data between hosts holding stored data and clients running applications making use of these data;
 ii.  Image storage formats such as DICOM, TIFF or JPEG, specifying how images should be encoded for long-term storage or transmission;
iii.  Image data compression standards, specifying standard methods for compressing image (and video) data for efficient transmission;
 iv.  Database command languages, providing a standard syntax for specifying queries to a relational database; and
  v.  Metadata standards, providing a framework for describing the content of multimedia objects, and languages such as XML in which to write content descriptions.

The Moving Picture Experts Group (MPEG) has been developing a standard for data description and representation specifically geared to the needs of CBIR (ISO/IEC 23004-1:2007, 2007; ISO/IEC 29116-1:2008, 2008; ISO/IEC 23008-2:2013; Ozer, 2013; Richardson, 2010). Almost every aspect of CBIR activity suffers the impact of norms like MPEG-7 (Beach, 2008) and MPEG 21 (ISO/IEC 21000, 2013; ISO/IEC 23008-2:2013; Akramullah, 2014).

The commonest image format for both storage and transfer is the Digital Imaging and Communications in Medicine (DICOM) (Mustra et al., 2008; Konig, 2005). Non-visual data sources may be included using formats like PDF for transfer purposes, JPEG 2000, JPIP combined in DICOM. Additionally, the IHE Cross-Enterprise Document Sharing for Imaging (XDS) is an industry standard for managing shared documents among healthcare enterprises (Schimdt et al., 2014). Representative healthcare applications for JPIP streaming comprise teleradiology, Health Information Exchange (HIE), Electronic Medical Records (EMR) or Electronic Health Records (HER), and Personal Health Record (PHR).

XML is a modern approach to the interoperable exchange of multimedia metadata between broad ranges of devices. Many XML documents do not fully utilize all the information present in a given schema; thus, users download substantial redundant information for the current application. The structure of an XML documents is often expressed using an XML schema which allows applications to check and validate instances of documents (Davies, 2006). However, there are wordiness problems: a single XML document can often link to several, potentially large, schemata. While it is not always necessary to possess the schema to receive an XML document, a valid modification of the XML requires available schemata (Davis & Burnett, 2005a; Davis & Burnett, 2005b). Many descriptors are available for multimedia content, and, consequently, many XML schemata have been generated. For example, there are schemata representing albums, pictures, music, audio descriptors, and movies, etc.. Some descriptors have been standardized in MPEG-7 (Manjunath, 2002; Pattanaik & Balke, 2012) and MPEG-21 (Davis, & Burnett, 2005a; Bormans, & Hill, 2002; Karpouzis et al., 2005).

Besides the content itself, there is an increasing usage of metadata to let the community label the content according to shared schemata. Whether the community chooses standardized descriptors (i.e., MPEG-7) or their creations, a large number of XML documents and schema documents must be exchanged between the community devices. These large, redundant descriptor sets are a significant problem in mobile environments. For communities to operate efficiently, all users need to be able to contribute to the growth of data, and hence editing of the descriptors is an important part of the process. The size of schemata was recognized as a problem by the MPEG community while standardizing MPEG-7, thus leading to MPEG-7 TEM and BiM (Bormans, & Hill, 2002). However, this mechanism is of little use in a collaborative environment where users must have the ability to access fragments of the schema as XML documents are parsed and processed.

To allow mobile device users to exploit XML without these penalties, an extended RXEP protocol can be employed to allow fragments of schemata to be delivered with requested pieces of XML (Davis, & Burnett, 2005b). So, users may create rich and vast sets of multimedia data descriptors with the knowledge that their schemata can be used in both high and low bandwidth environments.

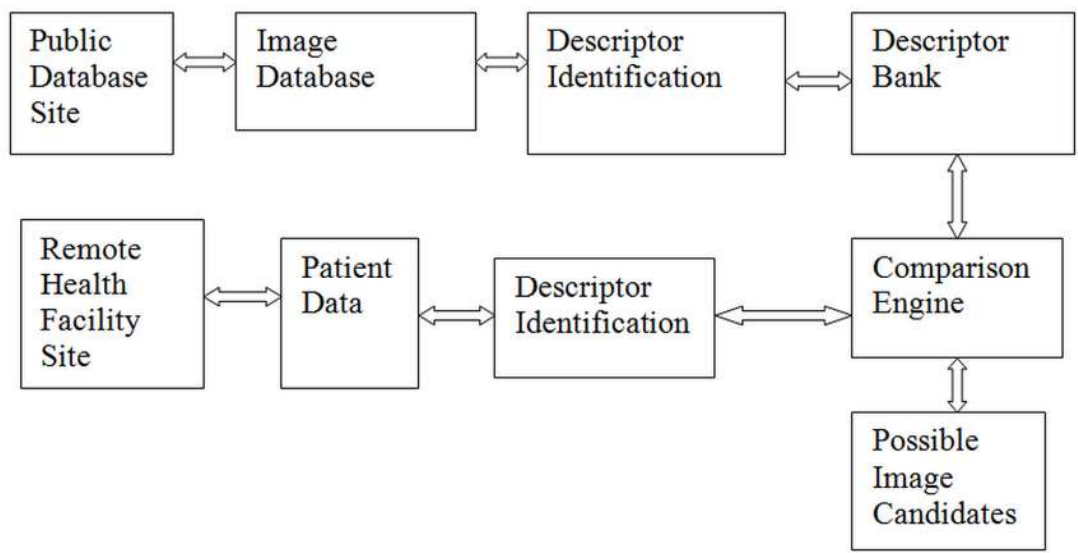

Figure 1. A typical CBIR system (CBIRS) architecture

## 3 CBIR SYSTEM ARCHITECTURE AND TECHNIQUES

A common approach is to incorporate computer vision knowledge to visual indexing. Each query image is analyzed to select features from a broad scope such as color, shape or other characteristic. The best feature set becomes an automated account of the image stored in a database, so that pictures can be retrieved by using the associated image features as an index.

The design of a CBIRS should mostly hinge on the search tasks that a user needs or wishes to perform. Questions regarding whether or not the user already knows the image sought after and its relevance for the application in mind play significant role. It might even occasionally be indispensable to retrieve images somewhat different from the desired one. Both manually inserted descriptive metadata and CBIR approaches are potential solutions to technical problems rather than design concerns. Figure 1 presents a CBIRS architecture that can be used to understand better the forthcoming explanations.

Next, some criteria are needed to compare visual data. The process of proposing selection conditions to filter images is named querying. A query such as '*give me all text files about lethal food*' can help developing procedures to seek out images in bulk in CBIRSs. As said before, content denotes shapes, class, colors, textures, or some extra particulars resulting from the picture itself. The concept of descriptor is also related to the localization of possible matches, since it is a short account of the facts looked for.

A CBIRS retrieves images with characteristics analogous to others under consideration and, at times, several pathologies are visually similar to the query wound, but belong to different diagnostic classes. A CBIRS allows one to mark retrieved images as being of positive and negative importance response (Datta et al., 2008) to improve diagnosis performance and, consequently, to better tune parameters. Appropriate CBIR web services can be used remotely to perform query-by-example in various image databases around the world and complement text-based retrieval methods. The system locates, retrieves and displays images similar in appearance to the one sought, using a set of features not linked to any particular investigative method able to visually typify the image. Intuitive and high-quality support to both inexpert and skilled users can improve diagnostic accuracy. The large diffusion of digital image archiving systems in medical institutions increases research interest to develop novel schemes for more efficient utilization of the vast amounts of health-related data (Jain, & Niranjan, 2008). Some elements have to be taken into account:

**1.** A pragmatic imaging protocol to normalize acquired images from various sources (ultrasound, PET, MRI, CT, SPECT, etc.) and that covers image storage, processing, and evaluation. The standards developed to categorize images still face scaling and miscategorization issues.

**2.** Image repositories need quality control and tools for data storage, distribution, and analysis. The archive access method is paramount to keep the response time low, even with huge files. Metadata-based systems have limitations as well as a significant range of potential applications which may degrade image retrieval efficiency. Textual data related to images can be effortlessly searched via existing technology, but this requires humans themselves to describe each database image, which can be impractical for big databases or when images are generated in real time.

**3.** Clinical facilities that engage patients to engender the images and to send them back to the imaging centers to optimize the use healthcare staff.

As CBIRSs require the definition of image descriptors, feature extraction, and feature matching of medical images, they can be very demanding due to the required resolution, databases sizes and search procedures. Query by example compares an input image to a database. The basic search algorithms may differ depending on the application, nevertheless resulting images should all share common traits with the example. Content comparison by means of distance measures aids decisions during the steps of locating, retrieving and displaying significant precedent cases with diagnostic reports. Particularly challenging aspects in this domain are to pull out local spots in specific image features and to set the similarity model. The most frequent image comparison approach in CBIR is via an image distance metric or similarity measure to match images in a variety of dimensions or categories, like shape, color, texture, significance and others (Datta et al., 2008). The notion of dimension refers to the number of axes used to represent an image (2D, 3D, 4D, etc.), so that distance or similarity metrics can be calculated. Blending specific criteria yields the best metrics, allowing the full ranking of the database with respect to the user submitted image. Category denotes a conception that does not require quantization, like shape, color, and meaning. Next, the main points for designing a good CBIRS are summarized.

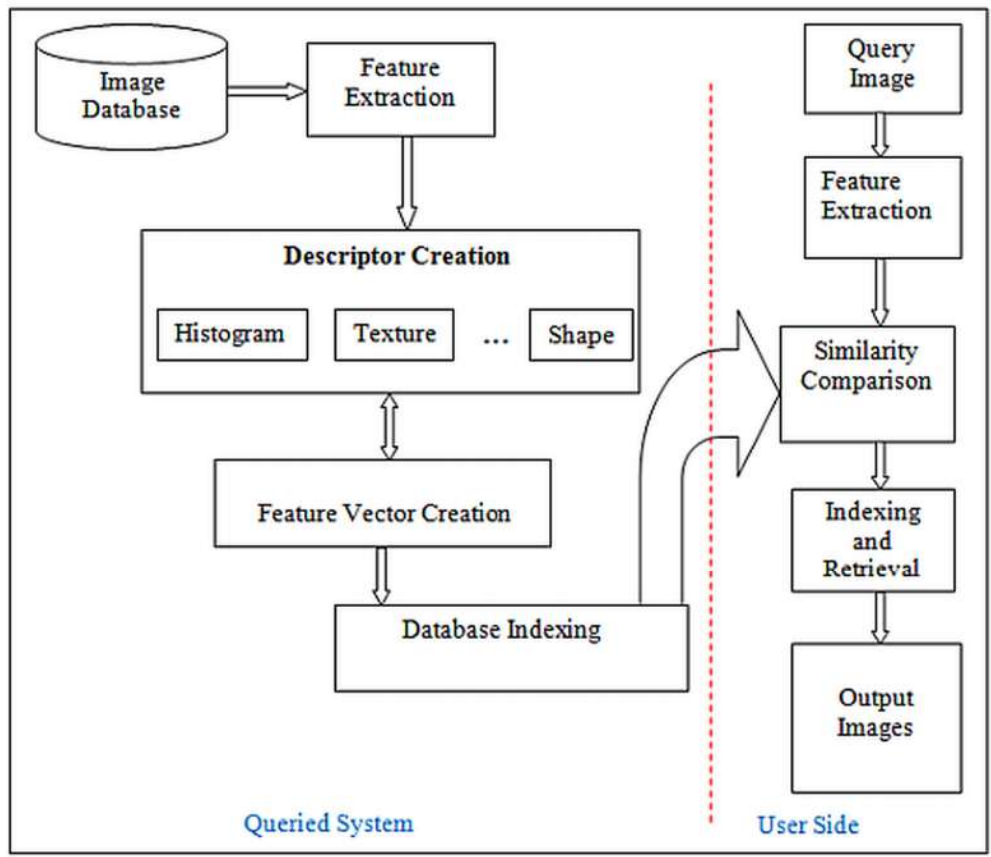

Figure 2. CBIRS from the functional point of view of the user and the queried system

## Steps to Develop a Good CBIR System (CBIRS):

1. Identify requisites and requirements of the needed CBIR system architecture.

2. Study the existing images retrieval methods and their efficacy.

3. Maintain a database of real-world medical cases and pictures collected from various sources and classifying them based upon the information provided by them.

4. Identify prospective visual features to mine and choose a similarity metric.

5. Web deployable structure to make it available to potential users via the Internet.

6. Limit access via authorization practices, with anonymity and security of patient data.

7. Efficient web-based GUIs, which allow the user to apply permissions, provide visual queries, investigate retrieved results, and so forth.

Successful CBIR deployment in healthcare domains relies on the capacity to handle user intentions. Relevance feedback can gradually refine the search outcomes by labeling images as "*relevant*", "*not relevant*", or "*neutral*". Later, the investigated query can be searched again with the new data. Searches depending on metadata rely on annotation quality and completeness. Manual registration via keywords or metadata in a bulky record can be time intensive and may fail to determine the keywords preferred to depict an image. The performance assessment of the keyword image exploration is subjective and has not been well-defined. Hence, the success of CBIR systems is difficult to measure (Chadha et al., 2012). It is also imperative to bear in mind web-based architectures, global availability, ease of access and inclusion of modern algorithms.

## 4 MAIN FOCUS OF THE CHAPTER
### 4.1 Feature Extraction
Figure 2 illustrates the processing stages involved in a CBIRS.

Feature Extraction can be done by text-based features and/or visual features. In the first case, keywords and annotations are used to build to portray and to index images. As for the second, one can rely on general attributes such as color, texture, shape and other domain-specific features that can be extracted (Huangyuan et al., 2014; Guo, & Kim, 2011; Grois et al., 2013; Grecos, & Wang, 2011). Wide-ranging features from the query and other images stored in the database are extracted, based on their pixel characteristics, so that the CBIRS stores image information in compact form into a separate database known as feature database, which is also known as image signature (Dhobale et al., 2011; Wong, 2002).

### 4.2 Image Description
Descriptors are the first step to finding out the connection between the pixels contained in a digital image and what humans call to mind after having observed an image or a group of images after some minutes. Visual descriptors are divided into two main groups: **a) General information descriptors** contain low-level descriptors that represent color, shape, regions, textures, motion, etc.; and **b) Specific domain information descriptors** provide information about objects and events in the scene. Next, several types of descriptors are discussed.

One way of organizing image descriptors is the way they describe content in terms of locality: global or local descriptors. Next, the most common content descriptors are explained. One should note that they can be combined to form a Feature Vector (FV).

**A. Color**
Color is an important visual factor. Based on the different applications different color spaces are available such as RGB (Red, Green, and Blue), LAB or CIELAB (L stands for lightness; **A** and **B** are the color opponent dimensions, based on nonlinearly compressed CIE XYZ color space coordinates), LUV, HSV (hue, saturation and value); HSL ( (hue, saturation, and lightness)), $YC_BC_R$ (family of color spaces used as a part of the color image pipeline in video and digital photography systems) and the hue-min-max-difference (HMMD). Despite the diversity of color models, the HSV has been deemed as the most appropriate due to its good perceptual nature (Mueller et al., 2004).

The opponent color axes model (R-G, 2B-R-G, R+G+B) has the advantage of isolating the brightness information on the third axis and it is invariant to changes in illumination intensity and shadows.

The HSV-representation is invariant under the orientation of the object with respect to the illumination and camera direction.

Search for clusters in a color histogram to identify which pixels in the image originate from one uniformly colored object. In image retrieval system, a color histogram related to a query image is calculated, then compared to the color histograms of other images stored in the database, and afterward, images are retrieved whose color histograms match those of the query most closely. A color histogram is represented as the proportion of pixels of each color represented in the color (Goel, & Sehgal, 2012). At the query time, the user can either indicate the color proportion, such as "blue 51%" or stipulate an example picture from which the color histogram is calculated. Histogram intersection was proposed in the work of (Swain & Ballard, 1991) as a feature matching option. Two color moments technique to replace the color histogram and reduced the quantization effects were devised by (Stricker & Orengo, 1995), and they focused on the thought that the color distribution can be typified by its moments. Color sets (Smith & Chang, 1996) were proposed as an estimate to the color histogram, to put into practice, it they transformed the RGB color space into a perceptually uniform space, such as HSV, and next, quantized the transformed color space into $M$ bins (Chada et al., 2012; Rui, Huang, & Chang, 1999).

The statistical moments is considered to be invariant to image shift, rotation and scale. Moments also represent fundamental geometric properties of a distribution of random variables.

**B. Texture**
Texture means visual patterns having homogeneity property and these cannot result from a single color or intensity. Examples of such surfaces are clouds, trees, bricks, hair, and fabric.

An approach for texture description based on psychological studies on human visual perception was proposed using computational approximations for six different visually meaningful texture properties, namely, coarseness, contrast, directionality, line-likeness, regularity, and roughness (Tamura et al., 1978). However, only three of the six proposed features correspond strongly to human perception and are widely used. These features are coarseness, contrast and directionality that describe, respectively, the Coarse × Fine, High × Low, and Directional × Non-directional of textured regions. In this proposal, we use these three described features in both descriptors.

Also, texture is an important quality to represent an image. Texture descriptors
characterize image textures or regions. They observe the region homogeneity and the histograms of these region borders. The set of descriptors can describe and embody, for instance, the following characteristics: Edge Histograms (EHs), Homogeneous Textures (HTs), and Texture Browsing (TB.)

Co-occurrence matrices can be constructed based on the orientation and distance between image pixels. Through the co-occurrence matrix, it is easy to calculate the contrast, coarseness, directionality and regularity, periodicity, directionality and randomness to understand the texture meaning. An approach that uses the statistics (mean and variance) extracted from the wavelet sub-bands as the texture representation can be found in (Smith, & Chang, 1996*)*. A technique using the wavelet transform with KL expansion and Kohonen maps to perform texture analysis can be found in (Gross et al., 1994). The wavelet transform was combined with a co-occurrence matrix in (Thyagarajan et al., 1994) to take advantage of both statistics-based and transform-based texture studies (Rui, Huang, & Chang, 1999).

**C. Shape**
Natural objects are primarily recognized by their shape. Some features related to object shape are computed for every object identified within each stored image. Shape representations can be divided into two categories, boundary-based and region-based. The former uses only the outer boundary of the shape while the latter uses the entire shape region (Smeulders et al., 2000). The most successful representation of these two categories is the use of Fourier descriptor and moment variants. The central idea of Fourier Descriptor is to use the Fourier transformed boundary as the shape feature. The main idea of Moment invariants is to use region-based moments, which are invariant to transformations as the shape feature. Queries to the system can be entered either in the form of example image or as a sketch (Rui, Huang, & Chang, 1999).

**D. Motion**
Contains valuable semantic information due to human's ability to recognize objects by their shape. However, this requires a segmentation similar to the one that the human visual system implements. Nowadays, such a segmentation system is not available yet. However there exists a serial of algorithms that are considered to be a good approximation. These descriptors describe regions, contours and shapes for 2D images and 3D volumes. The shape descriptors can represent, for instance, Region-based Shape (RS), Contour-based Shape (CS), and 3-D Shape (3-D S).

Motion is related to the objects motion in the sequence and the camera movement. This last information is provided by the capture device, whereas the rest is implemented using image processing. The descriptor set can describe Motion Activity (MA), Camera Motion Descriptor (CM), Motion Trajectory Descriptor (MT), and Warping and Parametric Motion (WM and PM).

**E. Location**
Elements location in the image is used to describe elements in the spatial domain. Also, elements can also be located in the temporal domain as Region Locator (RL) and Spatial-Temporal Locator (STL).

**F. Specific domain information**
They give information about objects and events in the scene. A concrete example would be face recognition. These descriptors, which give information about objects and events in the scene, are not readily extractable, even more when the extraction is to be automatically done. Nevertheless, they can be manually processed.

As mentioned before, face recognition is a concrete example of an application that tries to obtain automatically this information.

## 4.3 Feature Vector Creation
Once a set of descriptors is found, then a vector where its entries are features derived from the descriptors. If one choose to represent an image using 5 features, such as:
          Contrast [ T1 ]
          Dissimilarity [ T2 ]
          Homogeneity [ T3 ]
          Angular Second Moment [ T4 ]
          Entropy [ T5 ],

then a Feature Vector **FV** = [ T1,T2,T3,T4,T5] can represent a given image.

Next, a similarity metric (or metrics) can be selected to describe the FVs of the database images resembling the query image. For example, the Overall Similarity Metric (OSM):

$$OSM = (E_{texture} + E_{intenstiy} + E_{shape})/3, \qquad (1)$$

where $E_{texture}$, $E_{intenstiy}$, and $E_{shape}$ are partial similarity measures for texture, intensity and shape. Hence, the value of the OSM can be used to sort similar images, that is, past cases similar in some sense are retrieved. Similarity metrics and performance of comparison algorithms are investigated in the next Sub-sections.

### 4.4 Similarity Comparison

Image Similarity can be inferred from the image itself, such as difference of values of corresponding pixels and results from segmentation, and from the histogram analysis. Next some measures of similarity are discussed.

**Pixel Based Image Similarity**
Let **f** and **g** be two gray-value image functions. Hence,

$$pd1(\mathbf{a},\mathbf{b}) = \sum_{x=1}^{w}\sum_{y=1}^{h} |c(\mathbf{a},x,y) - c(\mathbf{b},x,y)|, \text{ and} \qquad (2)$$

$$pd2(\mathbf{a},\mathbf{b}) = \sum_{x=1}^{w}\sum_{y=1}^{h} (c(\mathbf{a},x,y) - c(\mathbf{b},x,y))^2. \qquad (3)$$

For example:

$$d\left(\begin{pmatrix} 4 & 3 & 7 \\ 0 & 0 & 1 \\ 9 & 5 & 5 \end{pmatrix}, \begin{pmatrix} 5 & 3 & 5 \\ 0 & 0 & 0 \\ 8 & 5 & 1 \end{pmatrix}\right) = \sqrt{1+4+1+1+16} = \sqrt{23} \quad . \qquad (4)$$

Let **a** and **b** be two images of size $w \times h$. Let $c$ be some image characteristic that assigns a number to each image pixels. As an example, $c(\mathbf{a},x,y)$ could be the gray value of the pixel at $(x, y)$.

**Pixel to pixel differences**
Differences between pixel characteristics can be represented by distinct metrics, such as:

$$pd1(\mathbf{a},\mathbf{b}) = \sum_{x=1}^{w}\sum_{y=1}^{h} |c(\mathbf{a},x,y) - c(\mathbf{b},x,y)|, \text{ and} \qquad (5)$$

$$pd2(\mathbf{a},\mathbf{b}) = \sum_{x=1}^{w}\sum_{y=1}^{h} (c(\mathbf{a},x,y) - c(\mathbf{b},x,y))^2. \qquad (6)$$

The statistical mean and variance can be used to add stability to pixel to pixel image difference:

$$spd(\mathbf{a},\mathbf{b}) = \sqrt{wh}\, \frac{\overline{d(\mathbf{a},\mathbf{b})}}{s_d(\mathbf{a},\mathbf{b})}, \text{ where} \qquad (7)$$

$$\overline{d(\mathbf{a},\mathbf{b})} = \frac{1}{wh}\sum_{x=1}^{w}\sum_{y=1}^{h} d(\mathbf{a},\mathbf{b},x,y), \tag{8}$$

$$s_d^2(\mathbf{a},\mathbf{b}) = \frac{1}{wh-1}\sum_{x=1}^{w}\sum_{y=1}^{h} (d(\mathbf{a},\mathbf{b},x,y) - \overline{d(\mathbf{a},\mathbf{b})})^2, \text{ and} \tag{9}$$

$$d(\mathbf{a},\mathbf{b},x,y) = c(\mathbf{a},x,y) - c(\mathbf{b},x,y). \tag{10}$$

Let $\mathbf{v(a)}$ be a vector of all $c(\mathbf{a},x,y)$ values assigned to all pixels in the image $\mathbf{a}$. Image similarity measures can be expressed as normalized inner products of such vectors. Since it yields maximum values for equal frames, a possible disparity measure is

$$vd(\mathbf{a},\mathbf{b}) = 1 - \frac{v(\mathbf{a}) \bullet v(\mathbf{b})}{\|v(\mathbf{a})\| * \|v(\mathbf{b})\|}. \tag{11}$$

### Image Histograms
The image histogram is a vector, so that, if $\mathbf{f}(x, y)$ is an image containing gray levels in the interval [0, 255], then $H(\mathbf{f})$ is its histogram, and $H(\mathbf{f})(k)$ is the number of pixels $(i, j)$, such that $F(i, j)=k$. Similar images $\mathbf{f}(.)$ and $\mathbf{g}(.)$ have similar histograms, but be aware that different images may have similar histograms:

$$d(\mathbf{H(f)},\mathbf{H(g)}) = (\Sigma_i[\mathbf{H(f)}(i) - \mathbf{H(g)}(i)]^2)^{-1/2}, \quad i \in [0, 255]. \tag{12}$$

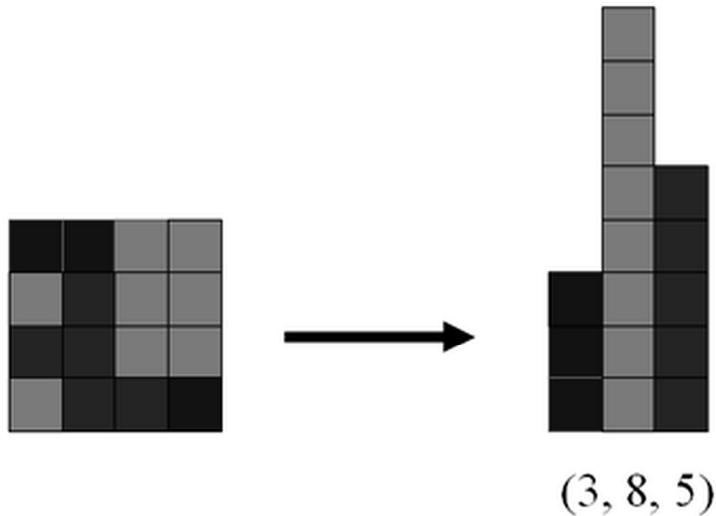

(3, 8, 5)

Figure 3. Image histogram for 3 gray levels. Image on the left and its histogram on the right

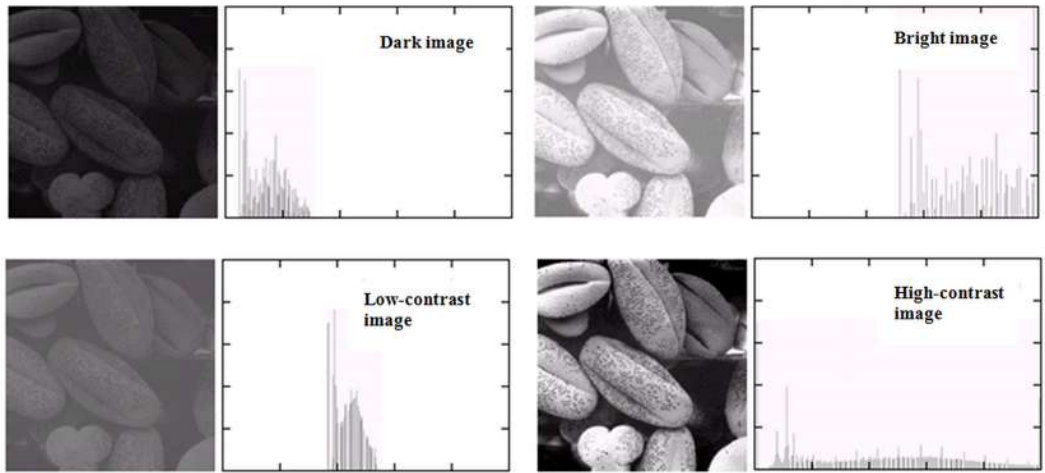

Figure 4. Relationship between some image properties and histograms

An example of a picture histogram can be seen in Figure 3. Figure 4 gives an idea of some properties that can be spotted by examining image histograms.

### 4.5 CBIR Evaluation

Humans can easily recognize the similarity between images. To test the effectiveness of a CBIRS, two evaluation measures, namely precision and recall are commonly used (Salton et al., 1975). Precision (*P*) is the ratio involving the number of relevant images retrieved (*NIR*), that is documents recovered by the system that are in fact relevant to the query, and the number of total retrieved images (*TID* as follows:

$$Precision = \frac{NIR}{TID}. \tag{13}$$

Recall (*R*) is the relative amount between the quantity of retrieved relevant images (*NIR*) and the total number of pertinent images existing in the database (*NID*):

$$Recall = \frac{NIR}{NID}. \tag{14}$$

High accuracy means that less pertinent images come from a query or that a more relevant range of images is recovered while high recall means few relevant images are neglected.

### 4.6 Indexing and Retrieving

In the matching procedure, the image signature of the query image is compared to the image signatures of the other images stored in the database. This involves the calculation of some distance measure between image signatures of the queried example and database images and, subsequently, to rank their images according to their distance threshold. Usually, the Euclidean distance is deemed as the most important metric. Next, by ranking, the system returns the outcomes that have a visual similarity to the query. Feature matching can be performed in two ways: (i) by region comparison, in which regions can be obtained by segmentation and, then, the distance between two regions is measured based on their low-level features; and (ii) by image comparison, which consists of number of regions. In recent years, several other distance measures have been developed for histograms such as city-block-distance and the Minkowsky distance (Khokher & Talwar, 2011).

## 4.7 Case Study: CBIR in Radiology

Radiology involves a typical application of CBIR in the healthcare domain. Some of the main issues involved in the design of such a CBIRS are:

i. Identification of potential users of the application.
ii. Study of the existing techniques for CBIR.
iii. Maintenance of a database of actual medical cases and images.
iv. Deciding upon which image features to be extracted and similarity metrics to be used.
v. Make the system web deployable.
vi. Organize authorized access to the CBIRS.
vii. Web-Based efficient Graphical User Interface (GUI)

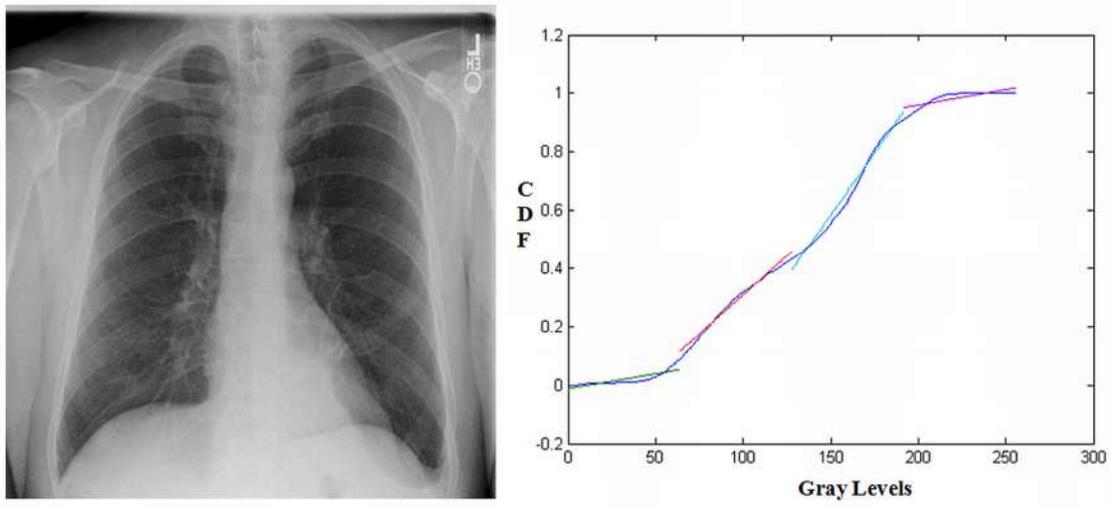

Figure 5. A radiography (left) and the corresponding CDF curve for $i \in [0, 255]$

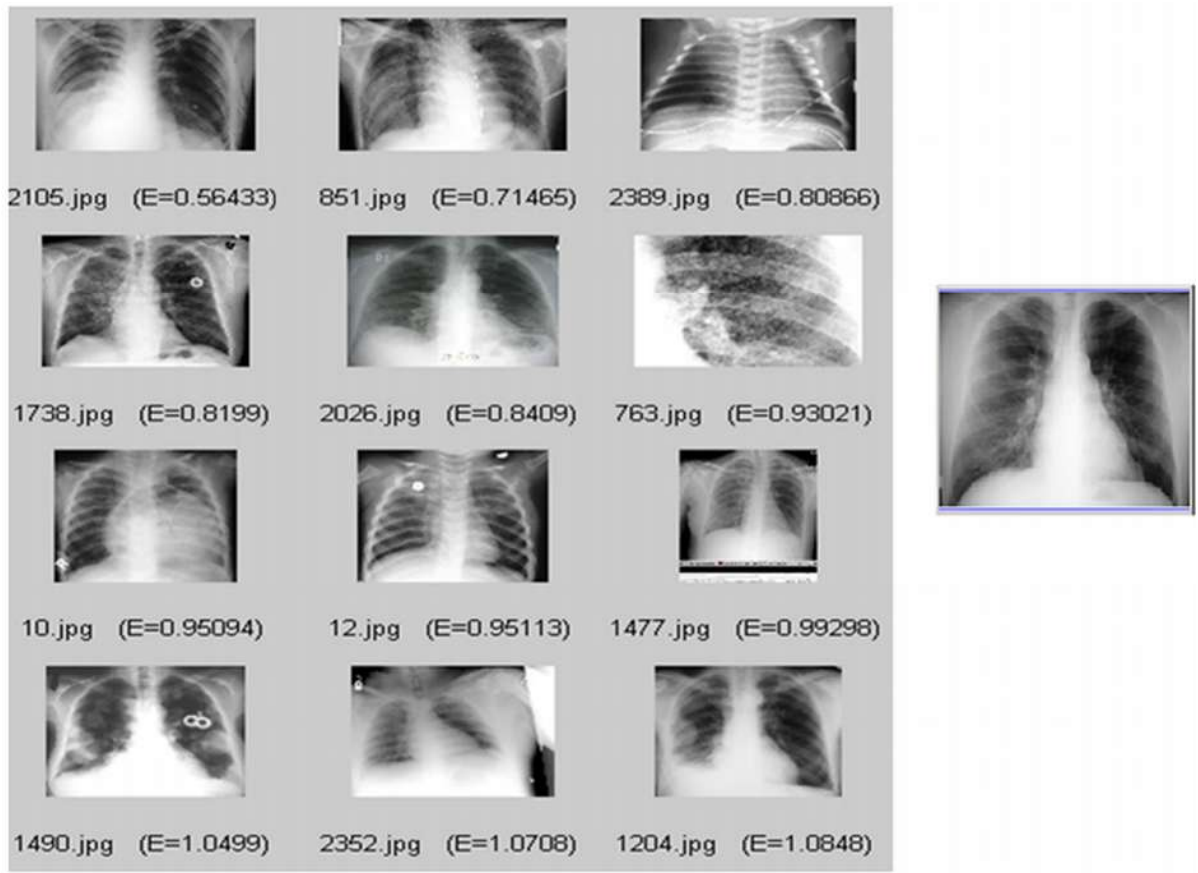

Figure 6. Possible images (left) corresponding to a given query image (right)

Figure 5 show the CDF plot for a radiological image. Such graph can be used to characterize it. Figure 6 present a set of images resulting from using the query image on the right side.

## 5 SOLUTIONS AND RECOMMENDATIONS

Development in the field of CBIR raises many research challenges, such as:

i. New image annotation techniques need to be developed because there are no such techniques available which properly deal with the semantic gap.
ii. CBIR system for specific domain/application needs to be developed by understanding their needs and information seeking behavior.
iii. Finding new connections, and mining patterns. Text mining techniques might be combined with visual-based descriptions.
iv. Matching query and stored images in a way that reflects human similarity judgments.

Probing big databases of images is a challenging endeavor, particularly, for retrieval by content. Most search engines compare the query image and all the database images and rank them by calculating their similarity measures or metrics, which are paramount to CBIR algorithms. These procedures search an image database to find images analogous to a certain query; therefore, they should be able to appraise the quantity of similarities between images.

The query image FV and the database image FV are compared using distance metrics or similarity measures. The images are ranked based on the distance values. Image retrieval with empirical evaluation has been used for a detailed comparison of different metrics in (Pentland et al., 1996).

As understood, the semantic gap is the lack of coincidence between the information that one can extract from the visual data and the interpretation that the same data have for a user in a given situation. A user seeks semantic similarity, but the database can only provide similarity by data processing, or there is an enormous amount of objects to be examined during search. Possible sources of errors:

   i. Incomplete query specification.
  ii. Incomplete image description.
 iii. Problems with color variances due to surface orientation, camera viewpoint position of illumination intensity of the light.
  iv. Problems with shape perception like occlusion and different points of view.
   v. Difficulties related to texture.
  vi. Object segmentation can cause setbacks.
 vii. Eventual existence of a narrow domain amid big data.

Lin et al. (2009) have used the difference between pixels of a scan pattern (DBPSP), the Color Co-occurrence Matrix (CCM), and the Color Histogram for K-Means (CHKM) as features in CBIRS. This algorithm decreases the image FV size, and the indexing time. Another technique for color image retrieval chooses as features color, texture, and shape information (Wang et al., 2010) and it delivers high retrieval effectiveness. Fu et al. (2006) have combined Gabor Filters (GFs) and Zernike Moments (ZMs) while considering texture and shape as features. GFs and ZMs are found helpful for face database.

A fuzzy logic structure (Xiu et al., 2003) has tried to alleviate problems in traditional CBIR systems, by considering the semantic gap and the perception subjectivity. Linguistic expressions provided a natural way of expressing the user's concepts, and membership functions characterized the mapping between image features and human visual concepts. It has also defined the syntax and semantics rules of a query description language to unify the query expression of textual descriptions, visual examples, and relevance feedbacks. The query comparison part inferred a similarity function based on the user's feedbacks to compare the query and each image in the database. The user's preference has been held, with the intention that his/her profile has been kept to achieve personalization. Experimental results showed that this structure lessened the semantic gap and the perception subjectivity problems.

Many fuzzy set methods have been employed to recover similar images. In (Li et al., 2003), a fuzzy similarity measure for CBIR has been employed to develop a faster similarity measure algorithm using the center of gravity of the fuzzy sets. An experimental CBIR system (Saha et al., 2004) has been developed using the Texture Co-occurrence Matrix.

Instead of global and local statistical features, a distinctive invariant feature set intended for CBIR has been proposed by (Wangming et al., 2008), where the visual contents of the query image and images are extracted from the database with an 128-dimensional FVs found by means of the Scale-Invariant Feature Transform (SIFT). Still, it would be desirable to reduce the dimension of the FVs to speed up retrieval. This work indexes and matches FVs combining the Best Bin First (BBF) KD-Tree, and the Approximate Nearest Neighbor (ANN) searching algorithms. Then, a voting scheme called Nearest Neighbor Distance Ratio Scoring (NNDRS) computes the aggregate scores of the corresponding candidate images in the database. By sorting the database images using total scores in decreasing order, the top few similar images can be revealed to users.

As the semantic gap between low-level and high-level semantics increases, the complexity of the retrieval task worsens (Murthy et al., 2010). The spotlight goes now from designing low-level image features to lessen the semantic gap between FVs and human semantics.

Relevance Feedback (RF) is a commonly used method to incorporate user's knowledge to the learning process for CBIR. Strategies for relevance feedback (Muller et al., 2000) in image retrieval to reduce the semantic gap were proposed. RF has been used for supervised learning with significant augment in the retrieval accuracy. Nevertheless, as a CBIR system continues to receive queries and feedbacks from the users, the data corresponding to their likings across query sessions are frequently lost at the end of search. This requires restarting of the feedback process for a new query. Some efforts targeting long-term learning have been made in general CBIR domain to alleviate this problem. Still, none addressed the needs and long-term similarity learning techniques for region-based image retrieval (Zhuo et al., 2014).

RF CBIRSs are considered in (Rudinac et al., 2007), where the influence of the type and the number of FV components on the retrieval efficiency has been investigated. They compared a CBIRS with a very small number of FV components (only 25 for color and texture) with a high-dimensional FV system inspired by MPEG-7.

The dimensionality curse problem appears when a large number of features reduces the image distinguishing ability and result in heavy indexing structures (Burges, 2005; Cheng et al., 2013; Hinton & Salakhutdinov, 2006). Consequently, it is imperative to maintain the dimensionality of the FV as lower as feasible. Dimensionality reduction (DR) techniques can help pos-process FVs, to choose the most important features, to throw away the irrelevant ones, and to turn the original feature space to a more representative one as it is done via a Self-Organizing Map (SOM)-based clustering method (Guo et al., 2008). This improves the overall speed and quality of the retrieval process. A limited set of relevant features simplifies the representation of visual knowledge; therefore, the necessary comparisons will be faster and utilize a smaller amount of storage. In other words, DR is the transformation of high-dimensional data into a meaningful representation of reduced dimensionality. The new intrinsic dimensionality of data is the lowest number of parameters required to account for the observed data properties. DR facilitates classification, visualization, and compression of high-dimensional data, by alleviating the dimensionality curse and other undesired properties of high-dimensional spaces. DR techniques can be linear and nonlinear. Linear methods presume that the data lie on or near a linear subspace of the high-dimensional space. Nonlinear DR techniques do not rely on the linearity assumption which leads to more complex embeddings of the data in the high-dimensional space. Performance investigations in medical image retrieval tasks have been done on artificial and natural data capitalizing on the findings from image processing (Pirolla et al.,2012; Verbeek, 2006; Wang et al., 2005), artificial intelligence (Bengio & LeCun, 2007), data mining ((Lee & Verleysen, 2005; Venkatarajan & Braun, 2004; Verbeek, 2006; Wang et al., 2005) and multimedia (Coelho & Estrela, 2012; Coelho & Estrela, 2013. Zhuo et al., 2014). Nonlinear techniques have performed better on selected artificial tasks, but do not surpass the more traditional methods like Principal Component Analysis (PCA) in real-world applications (Silva et al., 2011; Zhuo et al., 2014).

The use of a Support Vector Machine (SVM) classifier can improve retrieval rates (Silva et al., 2005; Cheng et al., 2013) and play a big role recently. Association rule mining is an emblematic method to mine data for revealing out of the ordinary trends, patterns and rules in huge data sets. Thus, SVMs and association rules are expected to be employed more to speed up image retrieval. Moreover, novel image classifying techniques such as Logitboost are still needed (Cat et al., 2006), because they may surpass SVMs.

The use of labeled images for learning is time-consuming. The Pseudo Labeling Method (PLM) (Wu et al., 2006) was proposed instead, using fuzzy rules to label images. To exploit the advantages of PLM and Fuzzy SVM (FSVM), an extended version of SVM with a unified framework called PLFSVM (Pseudo-Label FSVM) was designed.

The Latent Semantic Indexing (LSI) method (Chen et al., 2005) utilizes users' RF information. The intended retrieval system relies on region-based image and a Multiple Instance Learning (MIL) structure with One-Class Support Vector Machine (OSVM) as its core. MIL is a method to perform supervised learning (Zhang et al., 2005) with RF to guide the learning process.

## FUTURE RESEARCH DIRECTIONS

Retrieval systems based on low-level features are considered as unsatisfactory and unpredictable since these low-level features are not equivalent to human perception (Mamatha & Ananth, 2010). For such systems, it is hard to find adequate images based on a user query like "*Seek pictures having cysts and lumps*". Some researchers have tried to overpass the semantic gap by suggesting alternative techniques (Wang, Zhang, & Zhang, 2008).

Supervised machine intelligence groups a set of training images together and a binary classifier is taught to identify a semantic class label relying on some input metrics. Bayesian classifiers assign automatically database images to categories. Neural network is an additional method where the user chooses the classes (concepts) to be investigated: cell, organ, tissue, pathology, etc.. Next, a large amount of input data is given to the neural network classifiers to create the link connecting low-level features of an image and its high-level semantics (Veltkamp et al., 2008).

Unsupervised machine learning methods do not have outcome metrics, just a description of the input data organization and sets of images related to each other are associated with clusters. Each cluster is allotted some name. It maximizes the possibility of getting the similar image from that particular cluster (Veltkamp et al., 2008).

The RF technique is unfeasible in some domains. Its mechanism operates when the user enters the query in the form of an image, sketch or text. When the system retrieves related images from its database, the user checks the relevancy of the returned image. Then machine learning algorithm is applied to get the user feedback. This process is repeated till the user is satisfied with the results (Datta et al., 2008).

Web Image Retrieval techniques use the web information to retrieve images, such as the URL of a picture, image title, ALT-tag, descriptive text of the image, hyperlinks, etc.. But, its performance is not accurate. To improve its performance, research works have begun the use of visual image contents with the web information (Liu et al., 2006).

## CONCLUSION

The CBIR technology is yet unripe and used worldwide. This chapter provides a comprehensive survey of feature extraction techniques, image descriptors, similarity measures, semantic gap reduction techniques and performance measures. Although a significant amount of work has been done in this area, still there is no generic approach for high-level semantic-based image retrieval. To design a full-fledged image retrieval system with high-level semantics requires the integration of primitive feature extraction and high-level semantics extraction parameters. Open research issues are identified, and future research directions suggested.

CBIR is used to search a particular image from a large database, and to do interactive search of images from the database. Image retrieval does not entail solving the general image understanding problem. It may be sufficient that a retrieval system present similar images, similar in some user-defined sense. Other important issues are: interaction, the need for a well-designed database, and the semantic gap.

Feature extraction methods for CBIR comprise an imperative topic. Most recent CBIR techniques are designed to retrieve some image part, depending on the automatic extraction and comparison of image features considered most probably to express appearance. The features most often used include color, texture, shape, spatial layout, and multi-resolution pixel intensity transformations such as wavelets. This classification of feature set can be enhanced to heterogeneous (shape, texture) so that a more precise result can be obtained. It can also enhance the integration of heterogeneous features and the use of neural networks.

The multi-, trans- and interdisciplinary characters of the medical CBIR deployment emphasize the need for collaborative strategies as well as compatible standards. Due to the crescent number of technologies, convergence issues,

**ADDITIONAL READING SECTION**

**KEY TERMS & DEFINITIONS**

Content: It might refer to colors, shapes, and categories resembling the query, textures, or any other relevant clue that can result from the image collection on the basis of syntactical image features.

Content-Based Image Retrieval (CBIR): It is a framework that locates, retrieves and displays images alike to one given as a query, using a set of features and image descriptors.

Feature Vector (FV): It is a vector that contains numbers where each one represents an image characteristic or metric.

Image Processing (IP): Set of techniques and models to deal with image transformations in order to improve their quality and/or generate results based on input images. These results can be another image or some sort of knowledge extracted from an image set.

Image Descriptor (ID): Model and/or data structure used for describing an image.

Metadata: It comprises what is known about the image items contained in a large database, for example, who inserted the data and in which format. Therefore, metadata may refer to the knowledge that describes the context of another value. The words type, attribute, object, property, aspect, and schema all refer to metadata in some context.

Query Image: Image the user enters in order to obtain information.

Relevance Feedback (RF): It is a technique that involves the user interaction in the retrieval process by entering the query in the form of a image, sketch or text, but it is unfeasible in some domains. When the system retrieves related images from its database, the user checks the relevancy of the returned image according to some criteria.

Schema: It corresponds to the organization or structure of a database. The process of data modeling leads to a schema (whose plural is schemata).

Semantic Gap: It is the lack of coincidence between the data that one can extract from the visual information and the interpretation that the same data have for a user in a given situation.

Similarity Metric: It is a metric or distance employed to assess the quality of an image. In general, a CBIRS utilizes different similarity metrics.